\pdfoutput=1

\documentclass[11pt]{article}

\usepackage[final]{acl}

\usepackage{times}
\usepackage{latexsym}

\usepackage[T1]{fontenc}

\usepackage[utf8]{inputenc}

\usepackage{microtype}

\usepackage{inconsolata}

\usepackage{graphicx}

\usepackage{amsmath}

\title{Behavioural vs. Representational Systematicity in End-to-End Models:\\ An Opinionated Survey}

\author{
 \textbf{Ivan Vegner\textsuperscript{1,4}$^*$}\quad
 \textbf{Sydelle de Souza\textsuperscript{1,4}$^*$} \quad
 \textbf{Valentin Forch\textsuperscript{2}} \\
 \textbf{Martha Lewis\textsuperscript{3}$^\ddagger$} \quad
 \textbf{Leonidas A.A. Doumas\textsuperscript{4}$^\ddagger$}
\\\\
 \textsuperscript{1}School of Informatics, University of Edinburgh\\
 \textsuperscript{2}Chemnitz University of Technology\\
 \textsuperscript{3}Institute of Logic, Language, and Computation, University of Amsterdam\\
 \textsuperscript{4}School of Philosophy, Psychology \& Language Sciences, University of Edinburgh\\
}

\begin{document}
\maketitle

\def\thefootnote{*}\footnotetext{Joint first authors \qquad $\ddagger$ Joint senior authors}

\def\thefootnote{\arabic{footnote}}

\begin{abstract}
A core aspect of compositionality, systematicity is a desirable property in ML models as it enables strong generalization to novel contexts. This has led to numerous studies proposing benchmarks to assess systematic generalization, as well as models and training regimes designed to enhance it. Many of these efforts are framed as addressing the challenge posed by Fodor and Pylyshyn. However, while they argue for systematicity of \textit{representations}, existing benchmarks and models primarily focus on the systematicity of \textit{behaviour}.
We emphasize the crucial nature of this distinction.
Furthermore, building on \citeposs{hadleySystematicityConnectionistLanguage1994} taxonomy of systematic generalization, we analyze the extent to which behavioural systematicity is tested by key benchmarks in the literature across language and vision. Finally, we highlight ways of assessing systematicity of representations in ML models as practiced in the field of mechanistic interpretability.

\end{abstract}

\section{Introduction}

A core feature of human cognition is the ability to understand and generate novel combinations of known concepts in systematic ways. To illustrate, understanding (and producing) the utterance ``\textit{the octopus ate the fish}'' implies the ability to also understand (and produce) ``\textit{the fish ate the octopus}''. This \textit{systematicity}, often considered fundamental to the concept of compositionality, has become an increasingly important evaluation criterion for artificial intelligence systems. Recent years have seen a proliferation of papers proposing both benchmarks to test for systematicity \citep[e.g.,][]{lakeGeneralizationSystematicityCompositional2018, hupkesCompositionalityDecomposedHow2020, kimCOGSCompositionalGeneralization2020, wuReCOGSHowIncidental2023, kim2023imagine, okawaCompositionalAbilitiesEmerge2023} as well as new model architectures designed to exhibit systematic behaviour \citep[e.g.,][]{locatello2020object, alias2021neural, soulos2024compositional, assouel2024oc}. Many of these works frame their contributions as addressing the challenge\footnote{Although F\&P do not explicitly propose a challenge, it is referred to as such in \citet{fodorConnectionismProblemSystematicity1990}.} posed by \citet[][F\&P]{fodorConnectionismCognitiveArchitecture1988}: to explain how systematic language competencies can emerge without structured mental representations. However, there is a crucial distinction that is often overlooked. F\&P specifically argued for the necessity of systematic representations in cognitive systems, while modern benchmarks and evaluations typically focus solely on testing for the presence of systematic behaviour. This conflation has led to misunderstandings and conflicting claims about the systematic generalization capabilities of artificial neural networks. This is further compounded by black-box nature of our models.

This paper aims to disentangle these issues by separating behavioural and representational systematicity. We argue that meaningful progress toward human-like systematic generalization requires appropriately-scoped behavioural claims backed up by rigorous mechanistic interpretability.
We develop our argument in three parts. First, we discuss the historical development of the concept of systematicity.
Drawing on lessons from psychology, we highlight how the focus has gradually shifted from representations to behaviour, as well as the impact of this shift. Second, we analyse key benchmarks in the language\footnote{Specifically, at the syntactic and semantic levels.} and vision literature, assessing the type of systematicity they evaluate using \citeauthor{hadleySystematicityConnectionistLanguage1994}'s \citeyearpar{hadleySystematicityConnectionistLanguage1994} framework of weak, quasi-, and strong systematicity. Finally, we explore current approaches to evaluating systematicity of representations in end-to-end trained Transformer models through the lens of mechanistic interpretability.
Our review concludes with a set of recommendations for evaluating claims of systematicity in machine learning (ML) systems. It must be noted that works measuring compositionality of representations using custom architectures (e.g., \citet{andreasMeasuringCompositionalityRepresentation2019}; \citet{mccoy_tensor_2020}; \citet{liSystematicGeneralizationEmergent2022}), while relevant, are beyond the scope of this paper.

\paragraph{Related Work}

Compositionality has received a great deal of attention in ML. However, although one of the most frequently invoked definitions of compositionality is derived from the human capacity for systematicity, it has largely been sidelined as a secondary concern.

\citet{mccurdyCompositionalBehaviorNeural2024} survey the NLP research community, asking them to rate their agreement with a definition of ``compositional behaviour'' (CB) and a set of statements regarding the investigation thereof. They define CB as a model's tendency to produce a correct output for input $I$ given that it produces correct output for human-defined component parts of $I$. They further decouple their definition from any notion of learning, making their definition of CB agnostic to the dataset(s) from which it is learned. We believe that systematicity is indeterminable under this definition. Absent a clear definition of the component parts and the contexts in which they appear during training, it is impossible to judge a model’s ability to systematically recombine them. This impairs even behavioural evaluation of a model's generalization capabilities.

\citet{russinFregeChatGPTCompositionality2024} provide a comprehensive historical overview of compositionality as an object of inquiry. They also review current trends in endowing ML architectures with compositional mechanisms. However, their central question of ``can contemporary neural networks replicate the behavioral signatures of compositionality'' (p. 11) also focuses on behaviour rather than representation. They only briefly discuss systematicity and do not engage with \citeauthor{hadleySystematicityConnectionistLanguage1994}'s \citeyearpar{hadleySystematicityConnectionistLanguage1994} distinctions between levels of systematicity, which is central to our analysis. Nevertheless, they do call for mechanistic interpretability of model representations to reinforce conclusions in behavioural tests of compositionality.

There are some reviews which discuss representational systematicity, albeit in different terms.
\citet{wattenbergRelationalCompositionNeural2024} survey key mechanisms for enabling neural networks to represent relationships between features---a long-standing question in cognitive science, known as the binding problem \citep[see][for an overview]{feldmanNeuralBindingProblems2013}. They propose a number of directions for future work with respect to
interpreting how trained models may be representing (or approximating) relational binding.
\citet{greffBindingProblemArtificial2020} argue that contemporary neural networks lack human-level compositional generalization abilities precisely because they are unable to solve the binding problem.
The authors provide a comprehensive theoretical review of the connection between the binding problem and compositionality. They present an extensive survey of mechanisms which may implement binding and the properties which such mechanisms should possess. Conversely, \citet{pavlickSymbolsGroundingLarge2023} argues that existing LLMs can already encode symbols and undertake symbolic processing. Despite their differences, these papers all underscore the importance of mechanistic interpretability in establishing the presence of systematic representations in neural models.

\section{Compositionality via Systematicity}

Often attributed to \citet{fregeUberSinnUnd1892}, the principle of compositionality is frequently stated as: ``the meaning of a complex expression is a function of the meanings of its parts and of the way they are syntactically combined'' \citep{parteeCompositionality2004}. It is essential in formal semantics, serves as a foundational element in most logical formal languages\footnote{Although there are some exceptions, see for example, \citet{hintikkaLanguageGames1979}.}, and its influence extends across modern linguistic theories \citep{halvorsenMontaguesUniversalGrammar1979,gazdarGeneralizedPhraseStructure1985,steedmanCombinatoryCategorialGrammar2019,joshiTreeAdjoiningGrammars2005}.
In technical terms, this principle can be understood as a mathematical relationship---syntax and semantics function as algebras, with meaning assignment acting as a homomorphism that maps syntax to semantics \cite{janssenChapter7Compositionality1997}.

It is crucial to note that the principle of compositionality is ``extremely theory-dependent'' \citep{parteeCompositionality2004}. Without establishing a theory of syntax, semantics, and the interaction between them, the principle is severely underspecified and can be made trivial if not completely vacuous \citep{westerstahlMathematicalProofsVacuity1998, kazmiCompositionalityFormallyVacuous1998}. We echo \citet{janssenChapter7Compositionality1997}: ``the principle [of compositionality] should not be considered an empirically verifiable restriction, but a methodological principle that describes how a system for syntax and semantics should be designed.'' We now turn to F\&P's definition of compositionality, which puts systematicity in the driver's seat.

\subsection{Fodor and Pylyshyn's Compositionality} In order to establish a syntax and semantics of the language of thought, \citet[][F\&P]{fodorConnectionismCognitiveArchitecture1988} build their definition of compositionality from the more basic notion of systematicity. They define systematicity as an entailment between cognitive capacities: a human capable of understanding the statement $aRb$, where $a, b$ are objects and $R$ is a relation, is necessarily able to understand $bRa$.
For example, for the vast majority of language users, understanding the sentence ``\textit{John loves Mary}'' necessarily implies the ability to understand and produce the sentence ``\textit{Mary loves John}'', and vice versa. F\&P argue that this capacity is evidence for the existence of structure and structure-sensitive operations in the human brain. They assert that this property arises because such sentences are composed from the same constituents: ``insofar as a language is systematic, a lexical item must make approximately the same semantic contribution to each expression in which it occurs.'' F\&P admit relaxations of this principle in language, such as context-sensitive word definitions, syncategorematic compositions such as ``good [X]'' and idioms. However, they maintain that for most language users to be able to make systematic linguistic inferences, human cognitive representations must be systematic too.

Importantly, F\&P position their definition of compositionality---consisting of systematicity and the related properties of productivity and inferential coherence---as a property of the representational system (specifically of human mental representations).
For F\&P, composition is a syntactic (structural) operation which produces a semantic result. By complement, the semantics of individual linguistic units are amenable to performing structural operations over them. They assert that in representations without such structure, human-like compositionality is incredibly unlikely to emerge. Indeed, F\&P's challenge is to explain how systematicity can arise in a representational system which does not posit structural operations over symbols \citep{fodorConnectionismProblemSystematicity1990}.

\subsection{Hadley's Three Levels of Systematicity}

Based on the work of F\&P, \citet{hadleySystematicityConnectionistLanguage1994} outlines three progressive levels of systematicity in language learning.

\textit{Weak systematicity} is defined as the ability to process sentences that use familiar words in new combinations, but only if those words have appeared in identical syntactic positions before elsewhere in the training data. For instance, if a system is trained on sentences like ``\textit{John loves Mary},'' ``\textit{Jim loves John},'' and ``\textit{Mary loves Jim},'' it should be able to process ``\textit{Mary loves John}'' as a test case.

A system possesses \textit{quasi-systematicity} if it has weak systematicity under recursion. Quasi-systematic models can process novel sentences that contain embedded clauses, provided both the main sentence and the embedded clause are structurally isomorphic to various sentences in the training corpus. For any successfully processed complex sentence, the training data must include a simple sentence demonstrating the same word in the same syntactic position in which it appears in the embedded clause.

To exhibit \textit{strong systematicity}, the system must be able to process novel sentences, including those with embedded clauses, even when words appear in syntactic positions that they did not occupy in the training data in any simple or embedded clause. Hadley states that strong systematicity is closest to human systematic generalization, as the system can integrate its understanding of syntax and semantics to process words in entirely novel structures.

\subsection{Systematicity and Productivity}\label{para:productivity}
Closely connected to systematicity is the notion of linguistic productivity, defined as the ability to make "infinite use of finite means" \citep[][quoting \citet{humboldtLanguageDiversityHuman1836}]{chomskyAspectsTheorySyntax1965}.
To disambiguate the two, many have redefined productivity to refer specifically to a capacity for understanding and producing unbounded inputs and outputs \citep[e.g.,][]{fodorConnectionismCognitiveArchitecture1988, hupkesCompositionalityDecomposedHow2020}, or ``length generalization'' in modern parlance.

We conjecture that Hadley's weak systematicity will yield no productivity, because for a sentence of an unseen length there will necessarily be at least one word in the sentence in a novel syntactic position.
A quasi-systematic model could exhibit productivity for some, but not all, sentences given its ability to process some recursive clauses. Finally, strong systematicity yields full productivity.

\section{From Representation to Behaviour}
\label{sec:rep_vs_beh}

The core of the distinction between representational and behavioural systematicity lies in the classical distinction between competence and performance \citep[][p. 2]{chomskyAspectsTheorySyntax1965}. Representational systematicity is a competence, a cognitive mechanism which ideally manifests itself in performance, i.e., systematic behaviour, except as mitigated by other factors such as compute limitations, noise, errors, etc. Below, we lay out the implications of this distinction for ML research.

\subsection{Operationalization}

F\&P defined systematicity as a relation between cognitive competencies. Hadley reformulated their definition in terms of the properties that must be true of learning systems claiming to possess systematicity. In other words, he put forth a framework for the \textit{operationalisation} of systematicity.

Operationalisation is a common practice in fields like psychology, wherein unobservable properties of black-box systems must be reasoned about via proxy measures.
For example, to study an unobservable property like memory function, a psychologist may operationally define memory in terms of recall from a list. Operationalisation, though, is not without its potential pitfalls. For any operational definition $A$ of a concept $B$, demonstration of $A$ is at best only evidence for the presence of $B$, and the strength of the evidence depends on the strength of the connection between concept and operationalisation. Psychologists have long grappled with these issues, leading to the development of subfields like measurement theory and psychometrics \citep[e.g.,][]{luce1996ongoing, stevens1946theory}.

Consider the use of the relational match to sample task in comparative cognition \citep[e.g.,][]{premackDoesChimpanzeeHave1978}. In this task, the participant is presented with a pair of sample items (e.g., two red shapes) and asked to select one of two additional pairs of items that best corresponds to the sample (e.g., two blue shapes vs a red shape and a blue shape). The task aimed to test the capacity to reason about a relational concept like same/different, and was historically used as a demonstration that certain animals had or did not have this capacity \citep[e.g.,][]{thompsonLanguagenaiveChimpanzeesPan1997, fagotRelationalMatchingBaboons2010}. However, Mike Young and colleagues \cite{fagotDiscriminatingRelationRelations2001, youngEntropyDetectionPigeons1997, youngEntropyVariabilityDiscrimination2001} demonstrated that certain animals solved the task simply by detecting and responding to the entropy in sets of items, without recourse to relational reasoning.

The converse problem exists as well. The field of developmental psychology has long dealt with the issue of performance and competence with children, for whom failure to exhibit a behaviour does not necessarily imply the lack of capacity to do so in any circumstance (such as due to distraction or disinterest). In the literature on object permanence, for example, there has been a long debate about if and when children actually possess the capacity \citep[e.g.,][]{piagetConstructionRealityChild2013,baillargeonObjectPermanenceYoung1991}.
A resolution one way or the other would circumscribe the scope of what a child is able to do and understand at a given age. However, this has proven elusive due to the difficulty of operationalising object permanence appropriately.

\subsection{Implications for Machine Learning}

There are parallels to be drawn here with research in ML.
The literature is rife with failures of operationalisation, in which a model's performance on a particular task poorly predicts performance on other, even very similar tasks \citep[see][for an overview of ``shortcut learning'']{geirhosShortcutLearningDeep2020}. Thus, modellers must ensure that their tasks can only be solved using the competence which they intend to test.

However, demonstrating perfect operationalisation is very difficult. Consequently, when evaluating generalization capabilities, researchers require insights into the underlying representations that drive the behavioural outcomes on their task of choice. This \textit{mechanistic interpretability} enables researchers to assess the validity of their operationalisation and predict model performance across tasks that draw upon the same competence, even when behavioural tests prove imperfect.

According to F\&P, systematic generalization requires systematicity of internal representations
(\textit{representational systematicity}), not merely evidence of systematicity in the model's performance on any number of tasks (\textit{behavioural systematicity}). The relationship between valid operationalisation, behavioural evidence, and mechanistic interpretability creates three analytical cases, each with different implications for our understanding of model capabilities:

\paragraph{Case 1: Systematic behaviour without valid operationalisation.}
Valid operationalisation requires demonstrating that tasks actually demand specific levels of behavioural systematicity, rather than weaker levels or no systematicity at all. This demands principled control over dataset syntax and semantics. If this is not established, success or failure on the behavioural task no longer constitutes evidence of the existence or absence of representational systematicity. Models may rely on various non-systematic strategies to achieve performance, even when datasets intuitively appear to require systematicity. Sections \ref{sec:behavioural} and \ref{sec:evidence_against} examine several imprecise operationalisations of systematicity and demonstrate how this inhibits drawing meaningful conclusions about the systematic generalization capabilities of evaluated models.

\paragraph{Case 2: Systematic behaviour under valid operationalisation.}
When models demonstrate systematic behaviour under conditions of valid operationalisation, F\&P assert that they must possess systematic representations. If one gives credence to this position, supported by Hadley and other theorists of structured cognition, this case will be evidence of systematic internal representations.
However, definitively proving this requires mechanistic interpretability of the model representations underlying systematic behaviour. Demonstrating that a model exhibits systematic behaviour without possessing systematic representations would effectively overcome Fodor and Pylyshyn's challenge, representing a significant theoretical breakthrough.

\paragraph{Case 3: No systematic behaviour under valid operationalisation.}
This case also permits two competing interpretations. The model may fail to display systematic behaviour because it does not have systematic representations. Alternatively, the model may possess systematic representations, but be unable to use them to behave systematically.

The second possibility is analogous to the child who cannot, or does not want to, demonstrate object permanence in a particular context. This possibility becomes more worrisome as ML systems become more complex and modular, with components performing one subtask pipelined with components solving another. Even if one module has representations that support systematicity, the system will be unable to demonstrate systematicity on the task if a non-systematic module intervenes before the output layer. One example of models ending up in this case (partially) as a result of dataset construction is illustrated in \citet{wuReCOGSHowIncidental2023}, which we discuss below. As in the case above, distinguishing between these two interpretations requires mechanistic interpretability of the model's representations. Both are consistent with F\&P's challenge.

\section{Behavioural Systematicity}\label{sec:behavioural}

We propose that Hadley's taxonomy provides a robust framework for operationalising systematic generalization. It clearly defines the generalization capabilities of each level of systematicity, and establishes the requirements for testing it in a particular train/test data split. Below, we use it to examine representative datasets that test behavioural systematicity in end-to-end trained models. Rather than attempting an exhaustive review of this rapidly evolving field, we focus on key examples that demonstrate the critical importance of evaluating the level of systematicity required by a particular dataset. We hope that researchers will find this analysis useful and in turn apply it to their models and datasets.

\subsection{Linguistic Tasks}

\paragraph{SCAN, NACS} \citet{lakeGeneralizationSystematicityCompositional2018} introduce SCAN, a synthetic dataset for testing compositionality in text. It consists of all 20,910 command phrases generated from a non-recursive phrase-structure grammar of 13 words. Each command unambiguously maps onto a sequence of actions (for example, ``jump opposite left and walk thrice'' maps to ``LTURN LTURN JUMP WALK WALK WALK''). It aims to test compositionality via a judicious set of data splits: Split 1 is an iid (independent, identically distributed) split with up to 99\% of the training datapoints withheld randomly. Split 2 tests a form of productivity by testing the network on phrases which map to longer action (output) sequences than any that occur in training. Split 3 asks the network to parse compositions with a held-out word that it had only ever seen in a primitive context before (e.g., parsing "run twice and jump" having only seen "run" as a primitive, but having also seen "look twice and jump" before).

We suspect that Split 1 of SCAN tests Hadley's definition of weak systematicity. \citet{lakeGeneralizationSystematicityCompositional2018} state that with even 98\% of the training set for Split 1 withheld, every command in the test set that does not have a conjunction also appears in the training set an average of 8 times. Given that conjunction-less commands must also appear in conjunctions, it is likely that even 2\% of Split 1 exhibits all possible commands in all possible positions. Split 2 tests productivity and thus requires at least quasi-systematicity\footnote{The fact that the SCAN grammar is non-recursive does not mean the model cannot rely on recursion internally.}.
Finally, Split 3 aims to test strong systematicity sans recursion. However, \citet{bastingsJumpBetterConclusions2018} argue that good performance on SCAN can be achieved with simple modelling tricks that require neither structured representations nor systematicity, demonstrating the difficulty of inferring anything about a model's representations based on the model's behaviour on the task.

\paragraph{PCFG SET}
\citet{hupkesCompositionalityDecomposedHow2020} propose the PCFG SET dataset, an artificial translation task similar to SCAN. In this task, input sequences---generated by a probabilistic context-free grammar (PCFG) and corresponding to string edit operations---must be ``translated'' into output sequences representing the result of the sequence of operations. The authors propose several splits which aim to test various aspects of (their definition of) compositional generalization\footnote{
The authors also propose splits testing the useful properties of Substitutivity, Localism and Overgeneralization, which are out of the scope of the current work.
\citet{dankersParadoxCompositionalityNatural2022} extends this framework to look for evidence of these factors in a natural-language machine translation dataset, and highlighting the difficulty of rigorously evaluating compositionality in real-world data.
}:

    The \textbf{Systematicity split} tests the model's ability to produce the correct output for a sequence of operations \textit{a b}, if it has not seen this (sub)sequence in the training dataset, but has seen \textit{a} and \textit{b} in other contexts. This split appears to test Hadley's weak systematicity. Although, unlike in SCAN, the PCFG SET grammar is recursive and therefore the dataset is non-exhaustive, it is likely that most operations will individually have been seen in most of the positions in which they will occur in the test dataset. This is because the parameters for constraining the length and recursive depth of the generated input-output pairs are the same for the train and test sets.

    The \textbf{Productivity split} involves training the model on sequences containing up to 8 operations, and testing it on sequences containing $9+$ operations. As argued in Section \ref{para:productivity}, this split tests at least quasi-systematicity, and may contain some cases requiring strong systematic generalization.

\paragraph{COGS, ReCOGS, SLOG}
Proposed by \citet{kimCOGSCompositionalGeneralization2020}  and extended by \citet{wuReCOGSHowIncidental2023} and \citet{liSLOGStructuralGeneralization2023}, the COGS family of datasets examines systematicity in a linguistic semantic parsing context.
Directly specifying the meaning of a representation is challenging; instead, semantic parsing involves mapping sentences to logical forms (LFs). It does so under the assumption that the LFs can then be mapped to meanings more directly \citep{wuReCOGSHowIncidental2023}. The task involves parsing multiple semantically-equivalent expressions into a (more) canonical semantic representation concerned with binding the semantic roles of events and actions to particular entities. Here, we discuss the original COGS dataset \citep{kimCOGSCompositionalGeneralization2020}, but our arguments apply equally to its variants.

The dataset operationalises semantic parsing as a sequence-to-sequence problem. The model's task consists of predicting the LFs (as strings) for a large set of natural-language sentences. Systematicity is evaluated by testing the model on a generalization set, constructed by systematically altering the sentences in the training set. These manipulation include:

\begin{itemize}
\item
\textbf{Novel pairings of primitives and syntactic roles} (e.g., Subject $\rightarrow$ Object and vice versa)

\item
\textbf{Alternations of verb structure} (e.g., ``Charlie blessed Emma'' $\rightarrow$ ``Lina was blessed'')

\item
\textbf{Inferences based on verb class}, i.e., the semantic roles of a verb (e.g., ``The cobra helped the dog'' $\rightarrow$ ``The cobra froze'')

\item
\textbf{Deeper recursion} than seen in training

\item
\textbf{Object modifiers to subject modifiers} (e.g., "Noah ate the cake on the plate" $\rightarrow$ "The cake on the table burned")

\end{itemize}

To the extent that the first three cases, termed "lexical generalization" in the paper, evaluate novel recombinations of familiar items and structures, and the fourth and fifth, termed "structural generalization", test the ability to process novel structures, all test cases require strong systematicity.

The changes proposed by ReCOGS \citep{wuReCOGSHowIncidental2023} illustrate the importance of the distinction between behavioural and representational systematicity. Compared to COGS, ReCOGS (1) removes irrelevant output tokens, (2) isolates the effects of length vs. depth generalization and (3) mitigates spurious correlations between syntactic position and variable indexing. None of these transformations alter the meanings of the LFs and, therefore, the nature of the capabilities which COGS purports to test; nevertheless, ReCOGS evokes much better performance on structural generalization from the same models evaluated in \citet{kimCOGSCompositionalGeneralization2020}. Thus, it is hard to draw any conclusions as to the true systematic generalization capabilities of any model evaluated on COGS, as any model's success or failure on COGS may be due to spurious and unintended effects of dataset construction, rather than the model's capability for systematic representation  \citep[see][]{csordasDevilDetailSimple2021}. We note that SLOG \citep{liSLOGStructuralGeneralization2023} incorporates the proposals of ReCOGS and adds 17 new systematic generalization cases, providing a compelling behavioural benchmark for strong systematicity.

\subsection{Visual Tasks}
In contrast to linguistic tasks, compositionality and systematicity in vision have only recently re-emerged as a focus of research.
While perception is undoubtedly compositional---complex representations are made up of simpler percepts and relations between them---the nature of compositionality in perception is contested \citep[see][]{landeCompositionalityPerceptionFramework2024}. Furthermore, properties such as systematicity and productivity have not been clearly defined for visual perception and must not necessarily mirror their linguistic definitions \citep{cummins1996systematicity, landeCompositionalityPerceptionFramework2024, campThinkingMaps2007}. Lacking a strong theory of the syntax of images, it is unclear how Hadley's levels of systematicity apply to visual concepts
\citep[see][]{cavanaghLanguageVision2021, landeMentalStructures2021, quilty-dunnBestGameTown2023}. %
Multiple datasets address these challenges by testing compositionality in computer vision through three distinct approaches.

The first approach involves the \textbf{disentanglement of generative variables}. Here, the aim is to test whether a model can represent stimuli as a conjunction of several properties or variables which are conditionally-independent given the stimulus \cite{schottVisualRepresentationLearning2021}. Specifically, models are trained on compositionally constructed inputs to either directly predict the variables of the underlying generative model \citep{schottVisualRepresentationLearning2021}, classify composites of real-world images \citep{liaoDoesContinualLearning2023}, or, given an input image $I$, produce an image $I^{\prime}$ which only differs with respect to a minimal change in the generative variables (e.g., exchange of colors between two objects while preserving their shape, size, and position) \citep{kim2023imagine}. These datasets can only test weak systematicity, as they (a) lack a hierarchical structure, precluding them from incorporating recursion, which is necessary for quasi-systematicity, and (b) do not require the models to make use of concepts in novel contexts (e.g., colour always relates to a shape's interior, never its outline).

The second approach involves \textbf{vision-based abstract reasoning tasks} designed to evaluate conceptual understanding rather than mere perceptual feature composition.
Models must first infer the underlying invariants across images, and use this information to produce an appropriate response---e.g., identifying rule violations \citep{zerroug2022benchmark}, choosing the right image from a candidate set \citep{zhang2019raven, odouard2022evaluating}, or generating rule-based solutions \citep{chollet2019measure, moskvichev2023conceptarc, assouel2022object}. The RAVEN domain \citep{zhang2019raven} has no hierarchical structure, although \citet{zerroug2022benchmark} make use of hierarchical rule compositions. The open-ended ARC domain \cite{chollet2019measure} allows for complex task structures as well as novel applications of concepts, but there is no systematic description of its construction process. Thus, ARC may constitute a test for strong compositionality but cannot be judged on an objective basis. Derivative works \cite{moskvichev2023conceptarc, assouel2022object} give clearer task descriptions but still lack a rigorous control over the presentation of concepts in novel contexts, also making them unsuitable for testing strong systematicity.

A third approach examines \textbf{behavioural systematicity in vision-language models} without committing to a theory of compositionality in vision, leveraging linguistic theories of systematicity to test their text component. However, they do not exhaustively manipulate the syntax of the image caption. Instead, they approximate it by placing novel atoms and compounds in various syntactic slots in the caption and test whether the model's behaviour is sensitive to such manipulations \citep[e.g.,][]{johnsonCLEVRDiagnosticDataset2017, thrushWinogroundProbingVision2022, maCREPECanVisionLanguage2023, lewisDoesCLIPBind2024}. These approaches either test for weak systematicity \citep[e.g.,][]{johnsonCLEVRDiagnosticDataset2017, lewisDoesCLIPBind2024} as they use indivisible objects and do not always change the contexts in which concepts are used, or can only be applied to pre-trained models \citep[e.g.,][]{thrushWinogroundProbingVision2022, maCREPECanVisionLanguage2023}, making it impossible to judge the model's true generalization capabilities given unknown training exposure\footnote{For an exemplary discussion of the pitfalls of constructing a dataset testing only for systematicity, see \citet{diwan2022winoground}.}.

\subsection{Evaluating Learning Trajectories} Systematicity also relates to the speed of knowledge acquisition---a systematic learner should require training data that scales linearly with the number of concepts, whereas for a non-systematic learner this scaling should be exponential, assuming that the underlying grammar allows full permutations of concepts. A compositional representation should also allow faster skill acquisition as learning a new task will not require a reconfiguration of the encoding space. \citet{okawaCompositionalAbilitiesEmerge2023} examine convergence rates of diffusion models, finding that the number of training steps required to learn new concepts indeed scales exponentially with Hamming distance in conceptual space.
\citet{thomm2024limits} also find a super-linear scaling in an algorithmic task requiring Transformer models to map input strings to the outputs of compositional algorithms.

\section{Representational Systematicity}
To recap, investigating a model's representations allows us to reason about the model's performance on behavioural tests of systematicity, even in light of imperfect operationalisation\footnote{Although there is a wide range of neurosymbolic approaches that explicitly bake in systematic representations \citep[e.g.,][]{maoNeuroSymbolicConceptLearner2019,coecke_mathematical_2010, sen2022neuro, badreddine_logic_2022, doumasTheoryRelationLearning2022}, we only discuss to representational systematicity in end-to-end trained models.}.
Mechanistic interpretability provides evidence towards compositional, if not explicitly systematic, representations. This includes how predictions are built through a model \cite{nostalgebraist_interpreting_2020}; how information can be routed, copied, and deleted through a network \cite{elhage2021mathematical}; how a kind of compressed sensing can be used to allow models to represent efficiently \cite{elhage2022superposition}; and the tension between representational compression versus the flexibility of full compositionality \citep{olahDistributedRepresentationsComposition2023}. In examining trained models for systematic representations, most work takes the approach of ``concept discovery'', i.e., looking for the presence and use of a specific concept in a model's activations \cite{sharkeyOpenProblemsMechanistic2025}.

\subsection{Evidence for}
\citet{tenneyBERTRediscoversClassical2019,tenneyWhatYouLearn2019} uses linear probing to identify the presence of linguistic concepts that might be used in systematic representations. Having received much attention in recent years, this approach has been applied to the discovery of circuits responsible for algorithmic computation \citep{bricken2023monosemanticity, cunninghamSparseAutoencodersFind2023, parkLinearRepresentationHypothesis2024}.

\citet{liEmergentWorldRepresentations2024} and \citet{nandaEmergentLinearRepresentations2023} use probing to argue that OthelloGPT, a Transformer trained on the game Othello, learns a `world model'---a systematic representation of the rules and dynamics of the game that is used to plan the next move\footnote{However, cf. \citet{jylin04OthelloGPTLearnedBag2024}}.

\citet{toddFunctionVectorsLarge2024} and \citet{merulloLanguageModelsImplement2024} go further, extracting components (specifically, `function vectors') of a trained network that can be recombined in other contexts to produce predictable output \citep[cf.][]{opielkaAnalogicalReasoningLarge2025}.

Finally, \citet{fengHowLanguageModels2024} and \citet{fengMonitoringLatentWorld2024} discover `binding vectors' that sufficiently large LLMs use to bind entities. These can be used to encode propositions in linguistic tasks. Whilst their experiment does not test for systematicity (propositions are subject-verb-object, but verbs are such that subject and object cannot be swapped), it does propose a possible mechanism for systematicity in LLMs.

\subsection{Evidence Against}\label{sec:evidence_against}
Much of the evidence against systematic representations can be summed up in a critique of probing, both linear and non-linear. Specifically, the ability to decode a given property from a model's representations does not guarantee that the model causally relies on this property \citep{belinkovProbingClassifiersPromises2022, sharkeyOpenProblemsMechanistic2025}. Indeed, the property may be decoded from the representations even if it is uncorrelated to task performance or distributed in the input data as random noise \cite{ravichanderProbingProbingParadigm2021}.

\citet{vafaEvaluatingWorldModel2024} suggest that world models learned by Transformers on several tasks are ``far less coherent than they appear''. For example, they train a Transformer on a dataset of taxi journeys in Manhattan, requiring it to output a valid sequence of turns from an origin to a destination. They demonstrate that the Transformer does not learn a consistent world model despite excellent next-token prediction performance. The model learns a number of fictitious and impossible road connections; consequently, when the model is made to detour from its chosen route, the proportion of valid routes generated decreases dramatically.

Furthermore, \citet{aljaafariInterpretingTokenCompositionality2025} show that models do not represent words and phrases as systematic compositions of tokens. Instead, these representations are spread across attention heads and layers. Finally, even when (weakly) systematic representations can be detected, they are not always used.
\citet{kobayashi_when_2024} train a vanilla Transformer model on a task that involves systematic recombination of latent variables for a regression task. The model learns to generalize on in-distribution tasks, and the vector of underlying latent variables can be decoded from its activations. However, the model does not use this knowledge when faced with OOD tasks, and therefore fails. The Transformer is only able to systematically generalize to OOD tasks when augmented with an additional hypernetwork that explicitly takes the decoded latent vector as input.

Overall, there is mixed evidence for the use of systematic representations in Transformer models. While systematic representations can at times be extracted or elicited from the model, they also frequently resort to non-systematic tricks and shortcuts to achieve performance.
This may be due to the conflict between efficient data encoding and the development of fully systematic representations \citep{olahDistributedRepresentationsComposition2023}. Furthermore, even when models do learn apparently systematic representations,
it is difficult to ascertain to what extent they use them on OOD tasks.

\section{Takeaways for Machine Learning}
Compositionality is rapidly emerging as a key empirical factor of strong, robust and interpretable generalization behaviour in ML systems. However, the study of compositionality draws on a rich theoretical foundation spanning diverse approaches to syntax and semantics. As a result, two definitions of compositionality rooted in different frameworks may have little in common when applied to modelling and prediction. As the volume of research on compositionality in ML systems increases, it is crucial to maintain clear definitions of the phenomena in question. In this work, we have focused on \citeauthor{fodorConnectionismCognitiveArchitecture1988} definition, but other definitions are certainly possible. We urge researchers to engage with existing literature and situate their definition in the context of previous work, enabling the field to make fair comparisons amongst the multitude of datasets, methods and models.

While many researchers reference F\&P's seminal work, they frequently overlook the fact that F\&P define compositionality through systematicity, specifically systematicity of representations rather than behaviour. This is a crucial distinction, as behavioural systematicity does not necessitate the presence of systematic representations. A model might achieve strong performance on systematicity benchmarks through mechanisms that do not involve systematic internal representations, such as memorization or task-specific heuristics. There is evidence that this is already the case--- \citet{sunValidityEvaluationResults2023} show that different compositional generalization datasets rank the same modelling approaches differently while purporting to test the same capability. To address this issue, researchers should be explicit about their focus on behavioural systematicity, representational systematicity, or both. Those claiming to address F\&P's challenge must demonstrate the presence or absence of systematic representations, while those working on behavioural measures should justify why their operationalisation of systematicity is valid.

When examining systematicity through \citeauthor{hadleySystematicityConnectionistLanguage1994}'s framework of weak, quasi-, and strong systematicity, we find that many current benchmarks fall short of testing strong systematicity---the level that most closely approximates human systematic generalization capabilities. Many datasets inadvertently test only for weak or quasi-systematicity, limiting our ability to assess whether artificial systems can achieve human-like systematic processing. Future benchmark development should explicitly target strong systematicity, with careful consideration of the theoretical requirements for such evaluation. \citet{liSLOGStructuralGeneralization2023} is an example of such a benchmark. However, strong systematicity is difficult to test for in large-scale pre-trained models, as we are blind to what they see during training. Consequently, claims of strong behavioural systematicity may need to be limited to custom trained models.

Above all, the emergence of mechanistic interpretability techniques offers promising approaches for investigating representational systematicity. Recent work on function vectors and binding mechanisms suggests that some neural networks may indeed have mechanisms required for systematicity, though often in unexpected ways. However, many studies conflate the discovery of interpretable components with proof of systematic processing---the mere presence of decomposable representations does not guarantee they are used systematically, nor that the discovered mechanisms are causally responsible for the model's systematic behaviour.

In sum, it is representations, not task performance, that drive generalization. Until we develop principled methods for controlling training distributions and validating representational mechanisms, the quest for strongly systematic learners will remain stifled by the opacity of existing models. Despite the challenges, we believe that rigorous assessments of representational systematicity are well worth the effort.

\newpage

\section{Limitations}
This paper discusses systematicity, an important facet of compositionality, and how it is currently tested across ML literature. Limitations of the paper include the following:
\begin{itemize}
    \item We concentrate on clarifying the definition of one facet of compositionality, i.e. systematicity. In future work, more aspects of compositionality should be treated in the same manner.
    \item We have discussed widely used benchmarks in the literature to highlight the points we are making about common practices in ML, rather than an exhaustive summary of all literature.
    \item The benchmarks we have discussed tend to be somewhat based in English, and our analysis has focussed on monolingual English models. Multilingual models may have different generalization abilities.

    \item Our current analysis primarily assumes phrase structure grammar frameworks, but different theoretical perspectives on syntax may uncover additional dimensions of systematicity that warrant further investigation. Therefore, a full treatment of systematicity under alternative syntactic theories such as tree-adjoining grammars \citep{joshiTreeAdjoiningGrammars2005}, construction grammar \citep{Goldberg1995}, or combinatory categorial grammar \cite{steedmanCombinatoryCategorialGrammar2019} represents an important direction for future work.
\end{itemize}

\section{Ethics}
As a survey paper, there is limited ethical impact of the form of bias in datasets or environmental impact of training. As mentioned above, we do concentrate on benchmarks and tasks phrased in English, which may have an impact on the conclusions we draw. However, overall, we feel that our emphasis on thorough testing of claims of systematicity, and call for behavioural systematicity to be backed up by representational systematicity, should have a positive ethical effect in pushing for the development of more reliable and predictable models.

\section*{Acknowledgements}
IV, SdS, ML, and LD gratefully acknowledge travel and collaboration funds (RIS International Collaboration Grant) from Research Innovation Scotland. SdS, VF, and ML gratefully acknowledge the opportunity to build this collaboration at the SFI Working Group: Representations in Minds and Artificial Systems at the Santa Fe Institute. IV and SdS acknowledge that this work was supported in part by the UKRI Centre for Doctoral Training in Natural Language Processing, funded by the UKRI (grant EP/S022481/1) and the University of Edinburgh, School of Informatics and School of Philosophy, Psychology \& Language Sciences.

\bibliography{references, custom}

\begin{thebibliography}{99}
\providecommand{\natexlab}[1]{#1}

\bibitem[{Aljaafari et~al.(2025)Aljaafari, Carvalho, and Freitas}]{aljaafariInterpretingTokenCompositionality2025}
Nura Aljaafari, Danilo~S. Carvalho, and André Freitas. 2025.
\newblock \href {https://doi.org/10.48550/arXiv.2410.12924} {Interpreting token compositionality in {LLMs}: {A} robustness analysis}.
\newblock \emph{arXiv preprint}.
\newblock ArXiv:2410.12924 [cs] version: 2.

\bibitem[{Andreas(2019)}]{andreasMeasuringCompositionalityRepresentation2019}
Jacob Andreas. 2019.
\newblock \href {https://doi.org/10.48550/arXiv.1902.07181} {Measuring {Compositionality} in {Representation} {Learning}}.
\newblock \emph{arXiv preprint}.
\newblock ArXiv:1902.07181 [cs].

\bibitem[{Assouel et~al.(2024)Assouel, Astolfi, Bordes, Drozdzal, and Romero-Soriano}]{assouel2024oc}
Rim Assouel, Pietro Astolfi, Florian Bordes, Michal Drozdzal, and Adriana Romero-Soriano. 2024.
\newblock Oc-clip: Object-centric binding in contrastive language-image pretraining.
\newblock In \emph{NeurIPS 2024 Workshop on Compositional Learning: Perspectives, Methods, and Paths Forward}.

\bibitem[{Assouel et~al.(2022)Assouel, Rodriguez, Taslakian, Vazquez, and Bengio}]{assouel2022object}
Rim Assouel, Pau Rodriguez, Perouz Taslakian, David Vazquez, and Yoshua Bengio. 2022.
\newblock Object-centric compositional imagination for visual abstract reasoning.
\newblock In \emph{ICLR2022 Workshop on the Elements of Reasoning: Objects, Structure and Causality}.

\bibitem[{Badreddine et~al.(2022)Badreddine, Garcez, Serafini, and Spranger}]{badreddine_logic_2022}
Samy Badreddine, Artur~d'Avila Garcez, Luciano Serafini, and Michael Spranger. 2022.
\newblock \href {https://doi.org/10.1016/j.artint.2021.103649} {Logic {Tensor} {Networks}}.
\newblock \emph{Artificial Intelligence}, 303:103649.
\newblock ArXiv:2012.13635 [cs].

\bibitem[{Baillargeon and DeVos(1991)}]{baillargeonObjectPermanenceYoung1991}
Renée Baillargeon and Julie DeVos. 1991.
\newblock \href {https://doi.org/10.1111/j.1467-8624.1991.tb01602.x} {Object {Permanence} in {Young} {Infants}: {Further} {Evidence}}.
\newblock \emph{Child Development}, 62(6):1227--1246.
\newblock \_eprint: https://onlinelibrary.wiley.com/doi/pdf/10.1111/j.1467-8624.1991.tb01602.x.

\bibitem[{Bastings et~al.(2018)Bastings, Baroni, Weston, Cho, and Kiela}]{bastingsJumpBetterConclusions2018}
Jasmijn Bastings, Marco Baroni, Jason Weston, Kyunghyun Cho, and Douwe Kiela. 2018.
\newblock \href {https://doi.org/10.18653/v1/W18-5407} {Jump to better conclusions: {SCAN} both left and right}.
\newblock In \emph{Proceedings of the 2018 {EMNLP} {Workshop} {BlackboxNLP}: {Analyzing} and {Interpreting} {Neural} {Networks} for {NLP}}, pages 47--55, Brussels, Belgium. Association for Computational Linguistics.

\bibitem[{Belinkov(2022)}]{belinkovProbingClassifiersPromises2022}
Yonatan Belinkov. 2022.
\newblock \href {https://doi.org/10.1162/coli_a_00422} {Probing {Classifiers}: {Promises}, {Shortcomings}, and {Advances}}.
\newblock \emph{Computational Linguistics}, 48(1):207--219.

\bibitem[{Bricken et~al.(2023)Bricken, Templeton, Batson, Chen, Jermyn, Conerly, Turner, Anil, Denison, Askell, Lasenby, Wu, Kravec, Schiefer, Maxwell, Joseph, Hatfield-Dodds, Tamkin, Nguyen, McLean, Burke, Hume, Carter, Henighan, and Olah}]{bricken2023monosemanticity}
Trenton Bricken, Adly Templeton, Joshua Batson, Brian Chen, Adam Jermyn, Tom Conerly, Nick Turner, Cem Anil, Carson Denison, Amanda Askell, Robert Lasenby, Yifan Wu, Shauna Kravec, Nicholas Schiefer, Tim Maxwell, Nicholas Joseph, Zac Hatfield-Dodds, Alex Tamkin, Karina Nguyen, Brayden McLean, Josiah~E Burke, Tristan Hume, Shan Carter, Tom Henighan, and Christopher Olah. 2023.
\newblock Towards monosemanticity: Decomposing language models with dictionary learning.
\newblock \emph{Transformer Circuits Thread}.
\newblock Https://transformer-circuits.pub/2023/monosemantic-features/index.html.

\bibitem[{Camp(2007)}]{campThinkingMaps2007}
Elisabeth Camp. 2007.
\newblock \href {https://doi.org/10.1111/j.1520-8583.2007.00124.x} {Thinking with {Maps}}.
\newblock \emph{Philosophical Perspectives}, 21(1):145--182.
\newblock \_eprint: https://onlinelibrary.wiley.com/doi/pdf/10.1111/j.1520-8583.2007.00124.x.

\bibitem[{Cavanagh(2021)}]{cavanaghLanguageVision2021}
Patrick Cavanagh. 2021.
\newblock \href {https://doi.org/10.1177/0301006621991491} {The {Language} of {Vision}*}.
\newblock \emph{Perception}, 50(3):195--215.
\newblock Publisher: SAGE Publications Ltd STM.

\bibitem[{Chollet(2019)}]{chollet2019measure}
Fran{\c{c}}ois Chollet. 2019.
\newblock On the measure of intelligence.
\newblock \emph{arXiv preprint arXiv:1911.01547}.

\bibitem[{Chomsky(1965)}]{chomskyAspectsTheorySyntax1965}
Noam Chomsky. 1965.
\newblock \href {https://www.jstor.org/stable/j.ctt17kk81z} {\emph{Aspects of the {Theory} of {Syntax}}}, 50 edition.
\newblock The MIT Press.

\bibitem[{Coecke et~al.(2010)Coecke, Sadrzadeh, and Clark}]{coecke_mathematical_2010}
Bob Coecke, Mehrnoosh Sadrzadeh, and Stephen Clark. 2010.
\newblock \href {https://doi.org/10.48550/arXiv.1003.4394} {Mathematical {Foundations} for a {Compositional} {Distributional} {Model} of {Meaning}}.
\newblock \emph{arXiv preprint}.
\newblock ArXiv:1003.4394 [cs].

\bibitem[{Csordás et~al.(2021)Csordás, Irie, and Schmidhuber}]{csordasDevilDetailSimple2021}
Róbert Csordás, Kazuki Irie, and Juergen Schmidhuber. 2021.
\newblock \href {https://doi.org/10.18653/v1/2021.emnlp-main.49} {The {Devil} is in the {Detail}: {Simple} {Tricks} {Improve} {Systematic} {Generalization} of {Transformers}}.
\newblock In \emph{Proceedings of the 2021 {Conference} on {Empirical} {Methods} in {Natural} {Language} {Processing}}, pages 619--634, Online and Punta Cana, Dominican Republic. Association for Computational Linguistics.

\bibitem[{Cummins(1996)}]{cummins1996systematicity}
Robert Cummins. 1996.
\newblock Systematicity.
\newblock \emph{The Journal of Philosophy}, 93(12):591--614.

\bibitem[{Cunningham et~al.(2023)Cunningham, Ewart, Riggs, Huben, and Sharkey}]{cunninghamSparseAutoencodersFind2023}
Hoagy Cunningham, Aidan Ewart, Logan Riggs, Robert Huben, and Lee Sharkey. 2023.
\newblock \href {https://doi.org/10.48550/arXiv.2309.08600} {Sparse {Autoencoders} {Find} {Highly} {Interpretable} {Features} in {Language} {Models}}.
\newblock \emph{arXiv preprint}.
\newblock ArXiv:2309.08600 [cs].

\bibitem[{Dankers et~al.(2022)Dankers, Bruni, and Hupkes}]{dankersParadoxCompositionalityNatural2022}
Verna Dankers, Elia Bruni, and Dieuwke Hupkes. 2022.
\newblock \href {https://doi.org/10.48550/arXiv.2108.05885} {The paradox of the compositionality of natural language: a neural machine translation case study}.
\newblock \emph{arXiv preprint}.
\newblock ArXiv:2108.05885 [cs].

\bibitem[{Didolkar et~al.(2021)Didolkar, Goyal, Ke, Blundell, Beaudoin, Heess, Mozer, and Bengio}]{alias2021neural}
Aniket Didolkar, Anirudh Goyal, Nan~Rosemary Ke, Charles Blundell, Philippe Beaudoin, Nicolas Heess, Michael~C Mozer, and Yoshua Bengio. 2021.
\newblock Neural production systems.
\newblock \emph{Advances in Neural Information Processing Systems}, 34:25673--25687.

\bibitem[{Diwan et~al.(2022)Diwan, Berry, Choi, Harwath, and Mahowald}]{diwan2022winoground}
Anuj Diwan, Layne Berry, Eunsol Choi, David Harwath, and Kyle Mahowald. 2022.
\newblock Why is winoground hard? investigating failures in visuolinguistic compositionality.
\newblock \emph{arXiv preprint arXiv:2211.00768}.

\bibitem[{Doumas et~al.(2022)Doumas, Puebla, Martin, and Hummel}]{doumasTheoryRelationLearning2022}
Leonidas A.~A. Doumas, Guillermo Puebla, Andrea~E. Martin, and John~E. Hummel. 2022.
\newblock \href {https://doi.org/10.1037/rev0000346} {A theory of relation learning and cross-domain generalization}.
\newblock \emph{Psychological Review}, 129:999--1041.
\newblock Place: US Publisher: American Psychological Association.

\bibitem[{Elhage et~al.(2022)Elhage, Hume, Olsson, Schiefer, Henighan, Kravec, Hatfield-Dodds, Lasenby, Drain, Chen, Grosse, McCandlish, Kaplan, Amodei, Wattenberg, and Olah}]{elhage2022superposition}
Nelson Elhage, Tristan Hume, Catherine Olsson, Nicholas Schiefer, Tom Henighan, Shauna Kravec, Zac Hatfield-Dodds, Robert Lasenby, Dawn Drain, Carol Chen, Roger Grosse, Sam McCandlish, Jared Kaplan, Dario Amodei, Martin Wattenberg, and Christopher Olah. 2022.
\newblock Toy models of superposition.
\newblock \emph{Transformer Circuits Thread}.
\newblock Https://transformer-circuits.pub/2022/toy\_model/index.html.

\bibitem[{Elhage et~al.(2021)Elhage, Nanda, Olsson, Henighan, Joseph, Mann, Askell, Bai, Chen, Conerly, DasSarma, Drain, Ganguli, Hatfield-Dodds, Hernandez, Jones, Kernion, Lovitt, Ndousse, Amodei, Brown, Clark, Kaplan, McCandlish, and Olah}]{elhage2021mathematical}
Nelson Elhage, Neel Nanda, Catherine Olsson, Tom Henighan, Nicholas Joseph, Ben Mann, Amanda Askell, Yuntao Bai, Anna Chen, Tom Conerly, Nova DasSarma, Dawn Drain, Deep Ganguli, Zac Hatfield-Dodds, Danny Hernandez, Andy Jones, Jackson Kernion, Liane Lovitt, Kamal Ndousse, Dario Amodei, Tom Brown, Jack Clark, Jared Kaplan, Sam McCandlish, and Chris Olah. 2021.
\newblock A mathematical framework for transformer circuits.
\newblock \emph{Transformer Circuits Thread}.
\newblock Https://transformer-circuits.pub/2021/framework/index.html.

\bibitem[{Fagot and Parron(2010)}]{fagotRelationalMatchingBaboons2010}
Joël Fagot and Carole Parron. 2010.
\newblock \href {https://doi.org/10.1037/a0017169} {Relational matching in baboons ({Papio} papio) with reduced grouping requirements}.
\newblock \emph{Journal of Experimental Psychology: Animal Behavior Processes}, 36(2):184--193.
\newblock Place: US Publisher: American Psychological Association.

\bibitem[{Fagot et~al.(2001)Fagot, Wasserman, and Young}]{fagotDiscriminatingRelationRelations2001}
Joël Fagot, Edward~A. Wasserman, and Michael~E. Young. 2001.
\newblock \href {https://doi.org/10.1037/0097-7403.27.4.316} {Discriminating the relation between relations: {The} role of entropy in abstract conceptualization by baboons ({Papio} papio) and humans ({Homo} sapiens)}.
\newblock \emph{Journal of Experimental Psychology: Animal Behavior Processes}, 27(4):316--328.
\newblock Place: US Publisher: American Psychological Association.

\bibitem[{Feldman(2013)}]{feldmanNeuralBindingProblems2013}
Jerome Feldman. 2013.
\newblock \href {https://doi.org/10.1007/s11571-012-9219-8} {The neural binding problem(s)}.
\newblock \emph{Cognitive Neurodynamics}, 7(1):1--11.

\bibitem[{Feng et~al.(2024)Feng, Russell, and Steinhardt}]{fengMonitoringLatentWorld2024}
Jiahai Feng, Stuart Russell, and Jacob Steinhardt. 2024.
\newblock \href {https://openreview.net/forum?id=0yvZm2AjUr} {Monitoring {Latent} {World} {States} in {Language} {Models} with {Propositional} {Probes}}.

\bibitem[{Feng and Steinhardt(2024)}]{fengHowLanguageModels2024}
Jiahai Feng and Jacob Steinhardt. 2024.
\newblock \href {https://doi.org/10.48550/arXiv.2310.17191} {How do {Language} {Models} {Bind} {Entities} in {Context}?}
\newblock \emph{arXiv preprint}.
\newblock ArXiv:2310.17191 [cs].

\bibitem[{Fodor and McLaughlin(1990)}]{fodorConnectionismProblemSystematicity1990}
Jerry Fodor and Brian~P. McLaughlin. 1990.
\newblock \href {https://doi.org/10.1016/0010-0277(90)90014-B} {Connectionism and the problem of systematicity: {Why} {Smolensky}'s solution doesn't work}.
\newblock \emph{Cognition}, 35(2):183--204.

\bibitem[{Fodor and Pylyshyn(1988)}]{fodorConnectionismCognitiveArchitecture1988}
Jerry~A. Fodor and Zenon~W. Pylyshyn. 1988.
\newblock \href {https://doi.org/10.1016/0010-0277(88)90031-5} {Connectionism and cognitive architecture: {A} critical analysis}.
\newblock \emph{Cognition}, 28(1):3--71.

\bibitem[{Frege(1892)}]{fregeUberSinnUnd1892}
Gottlob Frege. 1892.
\newblock \emph{Über {Sinn} und {Bedeutung}}, 1. auflage edition.
\newblock Zeitschrift für {Philosophie} und philosophische {Kritik}, {Neue} {Folge}. Pfeffer, Leipzig.

\bibitem[{Gazdar et~al.(1985)Gazdar, Klein, Pullum, and Sag}]{gazdarGeneralizedPhraseStructure1985}
Gerald Gazdar, Evan Klein, Geoffrey~K. Pullum, and Ivan~A. Sag. 1985.
\newblock \href {https://www.hup.harvard.edu/books/9780674344563} {\emph{Generalized {Phrase} {Structure} {Grammar}}}.
\newblock Harvard University Press.

\bibitem[{Geirhos et~al.(2020)Geirhos, Jacobsen, Michaelis, Zemel, Brendel, Bethge, and Wichmann}]{geirhosShortcutLearningDeep2020}
Robert Geirhos, Jörn-Henrik Jacobsen, Claudio Michaelis, Richard Zemel, Wieland Brendel, Matthias Bethge, and Felix~A. Wichmann. 2020.
\newblock \href {https://doi.org/10.1038/s42256-020-00257-z} {Shortcut learning in deep neural networks}.
\newblock \emph{Nature Machine Intelligence}, 2(11):665--673.
\newblock Publisher: Nature Publishing Group.

\bibitem[{Goldberg(1995)}]{Goldberg1995}
Adele Goldberg. 1995.
\newblock \emph{Constructions: {A} construction grammar approach to argument structure}.
\newblock University of Chicago Press, Chicago ; London.
\newblock Series Title: Cognitive theory of language and culture.

\bibitem[{Greff et~al.(2020)Greff, Steenkiste, and Schmidhuber}]{greffBindingProblemArtificial2020}
Klaus Greff, Sjoerd~van Steenkiste, and Jürgen Schmidhuber. 2020.
\newblock \href {https://doi.org/10.48550/arXiv.2012.05208} {On the {Binding} {Problem} in {Artificial} {Neural} {Networks}}.
\newblock \emph{arXiv preprint}.
\newblock ArXiv:2012.05208 [cs].

\bibitem[{Hadley(1994)}]{hadleySystematicityConnectionistLanguage1994}
Robert~F. Hadley. 1994.
\newblock \href {https://doi.org/10.1111/j.1468-0017.1994.tb00225.x} {Systematicity in {Connectionist} {Language} {Learning}}.
\newblock \emph{Mind \& Language}, 9(3):247--272.
\newblock \_eprint: https://onlinelibrary.wiley.com/doi/pdf/10.1111/j.1468-0017.1994.tb00225.x.

\bibitem[{Halvorsen and Ladusaw(1979)}]{halvorsenMontaguesUniversalGrammar1979}
Per-Kristian Halvorsen and William~A. Ladusaw. 1979.
\newblock \href {https://doi.org/10.1007/BF00126510} {Montague's ‘universal grammar’: {An} introduction for the linguist}.
\newblock \emph{Linguistics and Philosophy}, 3(2):185--223.

\bibitem[{Hintikka(1979)}]{hintikkaLanguageGames1979}
Jaakko Hintikka. 1979.
\newblock \href {https://doi.org/10.1007/978-1-4020-4108-2_1} {Language-{Games}}.
\newblock In Esa Saarinen, editor, \emph{Game-{Theoretical} {Semantics}: {Essays} on {Semantics} by {Hintikka}, {Carlson}, {Peacocke}, {Rantala}, and {Saarinen}}, pages 1--26. Springer Netherlands, Dordrecht.

\bibitem[{Humboldt(1836)}]{humboldtLanguageDiversityHuman1836}
Wilhelm~von Humboldt. 1836.
\newblock \emph{On {Language}: {On} the {Diversity} of {Human} {Language} {Construction} and {Its} {Influence} on the {Mental} {Development} of the {Human} {Species}}.
\newblock Google-Books-ID: \_UODbGlD4WUC.

\bibitem[{Hupkes et~al.(2020)Hupkes, Dankers, Mul, and Bruni}]{hupkesCompositionalityDecomposedHow2020}
Dieuwke Hupkes, Verna Dankers, Mathijs Mul, and Elia Bruni. 2020.
\newblock \href {https://doi.org/10.48550/arXiv.1908.08351} {Compositionality decomposed: how do neural networks generalise?}
\newblock \emph{arXiv preprint}.
\newblock ArXiv:1908.08351.

\bibitem[{Janssen and Partee(1997)}]{janssenChapter7Compositionality1997}
Theo M.~V. Janssen and Barbara~H. Partee. 1997.
\newblock \href {https://doi.org/10.1016/B978-044481714-3/50011-4} {Chapter 7 - {Compositionality}}.
\newblock In Johan van Benthem and Alice ter Meulen, editors, \emph{Handbook of {Logic} and {Language}}, pages 417--473. North-Holland, Amsterdam.

\bibitem[{Johnson et~al.(2017)Johnson, Hariharan, Van Der~Maaten, Fei-Fei, Zitnick, and Girshick}]{johnsonCLEVRDiagnosticDataset2017}
Justin Johnson, Bharath Hariharan, Laurens Van Der~Maaten, Li~Fei-Fei, C.~Lawrence Zitnick, and Ross Girshick. 2017.
\newblock \href {https://doi.org/10.1109/CVPR.2017.215} {{CLEVR}: {A} {Diagnostic} {Dataset} for {Compositional} {Language} and {Elementary} {Visual} {Reasoning}}.
\newblock In \emph{2017 {IEEE} {Conference} on {Computer} {Vision} and {Pattern} {Recognition} ({CVPR})}, pages 1988--1997, Honolulu, HI. IEEE.

\bibitem[{Joshi(2005)}]{joshiTreeAdjoiningGrammars2005}
Aravind~K. Joshi. 2005.
\newblock \href {https://doi.org/10.1093/oxfordhb/9780199276349.013.0026} {Tree-{Adjoining} {Grammars}}.
\newblock In Ruslan Mitkov, editor, \emph{The {Oxford} {Handbook} of {Computational} {Linguistics}}, page~0. Oxford University Press.

\bibitem[{{jylin04} et~al.(2024){jylin04}, JackS, Karvonen, and {Can}}]{jylin04OthelloGPTLearnedBag2024}
{jylin04}, JackS, Adam Karvonen, and {Can}. 2024.
\newblock \href {https://www.alignmentforum.org/posts/gcpNuEZnxAPayaKBY/othellogpt-learned-a-bag-of-heuristics-1} {{OthelloGPT} learned a bag of heuristics}.

\bibitem[{Kazmi and Pelletier(1998)}]{kazmiCompositionalityFormallyVacuous1998}
Ali Kazmi and Francis~Jeffry Pelletier. 1998.
\newblock \href {https://www.jstor.org/stable/25001725} {Is {Compositionality} {Formally} {Vacuous}?}
\newblock \emph{Linguistics and Philosophy}, 21(6):629--633.
\newblock Publisher: Springer.

\bibitem[{Kim and Linzen(2020)}]{kimCOGSCompositionalGeneralization2020}
Najoung Kim and Tal Linzen. 2020.
\newblock \href {https://doi.org/10.18653/v1/2020.emnlp-main.731} {{COGS}: {A} {Compositional} {Generalization} {Challenge} {Based} on {Semantic} {Interpretation}}.
\newblock In \emph{Proceedings of the 2020 {Conference} on {Empirical} {Methods} in {Natural} {Language} {Processing} ({EMNLP})}, pages 9087--9105, Online. Association for Computational Linguistics.

\bibitem[{Kim et~al.(2023)Kim, Singh, Park, Gulcehre, and Ahn}]{kim2023imagine}
Yeongbin Kim, Gautam Singh, Junyeong Park, Caglar Gulcehre, and Sungjin Ahn. 2023.
\newblock Imagine the unseen world: a benchmark for systematic generalization in visual world models.
\newblock \emph{Advances in Neural Information Processing Systems}, 36:27880--27896.

\bibitem[{Kobayashi et~al.(2024)Kobayashi, Schug, Akram, Redhardt, Oswald, Pascanu, Lajoie, and Sacramento}]{kobayashi_when_2024}
Seijin Kobayashi, Simon Schug, Yassir Akram, Florian Redhardt, Johannes~von Oswald, Razvan Pascanu, Guillaume Lajoie, and João Sacramento. 2024.
\newblock \href {https://doi.org/10.48550/arXiv.2407.12275} {When can transformers compositionally generalize in-context?}
\newblock \emph{arXiv preprint}.
\newblock ArXiv:2407.12275 [cs].

\bibitem[{Lake and Baroni(2018)}]{lakeGeneralizationSystematicityCompositional2018}
Brenden Lake and Marco Baroni. 2018.
\newblock \href {https://proceedings.mlr.press/v80/lake18a.html} {Generalization without {Systematicity}: {On} the {Compositional} {Skills} of {Sequence}-to-{Sequence} {Recurrent} {Networks}}.
\newblock In \emph{Proceedings of the 35th {International} {Conference} on {Machine} {Learning}}, pages 2873--2882. PMLR.
\newblock ISSN: 2640-3498.

\bibitem[{Lande(2021)}]{landeMentalStructures2021}
Kevin~J. Lande. 2021.
\newblock \href {https://doi.org/10.1111/nous.12324} {Mental structures}.
\newblock \emph{Noûs}, 55(3):649--677.
\newblock \_eprint: https://onlinelibrary.wiley.com/doi/pdf/10.1111/nous.12324.

\bibitem[{Lande(2024)}]{landeCompositionalityPerceptionFramework2024}
Kevin~J. Lande. 2024.
\newblock \href {https://doi.org/10.1002/wcs.1691} {Compositionality in perception: {A} framework}.
\newblock \emph{WIREs Cognitive Science}, 15(6):e1691.
\newblock \_eprint: https://onlinelibrary.wiley.com/doi/pdf/10.1002/wcs.1691.

\bibitem[{Lewis et~al.(2024)Lewis, Nayak, Yu, Merullo, Yu, Bach, and Pavlick}]{lewisDoesCLIPBind2024}
Martha Lewis, Nihal Nayak, Peilin Yu, Jack Merullo, Qinan Yu, Stephen Bach, and Ellie Pavlick. 2024.
\newblock \href {https://aclanthology.org/2024.findings-eacl.101/} {Does {CLIP} {Bind} {Concepts}? {Probing} {Compositionality} in {Large} {Image} {Models}}.
\newblock In \emph{Findings of the {Association} for {Computational} {Linguistics}: {EACL} 2024}, pages 1487--1500, St. Julian's, Malta. Association for Computational Linguistics.

\bibitem[{Li et~al.(2023)Li, Donatelli, Koller, Linzen, Yao, and Kim}]{liSLOGStructuralGeneralization2023}
Bingzhi Li, Lucia Donatelli, Alexander Koller, Tal Linzen, Yuekun Yao, and Najoung Kim. 2023.
\newblock \href {https://doi.org/10.18653/v1/2023.emnlp-main.194} {{SLOG}: {A} {Structural} {Generalization} {Benchmark} for {Semantic} {Parsing}}.
\newblock In \emph{Proceedings of the 2023 {Conference} on {Empirical} {Methods} in {Natural} {Language} {Processing}}, pages 3213--3232, Singapore. Association for Computational Linguistics.

\bibitem[{Li et~al.(2024)Li, Hopkins, Bau, Viégas, Pfister, and Wattenberg}]{liEmergentWorldRepresentations2024}
Kenneth Li, Aspen~K. Hopkins, David Bau, Fernanda Viégas, Hanspeter Pfister, and Martin Wattenberg. 2024.
\newblock \href {https://doi.org/10.48550/arXiv.2210.13382} {Emergent {World} {Representations}: {Exploring} a {Sequence} {Model} {Trained} on a {Synthetic} {Task}}.
\newblock \emph{arXiv preprint}.
\newblock ArXiv:2210.13382 [cs].

\bibitem[{Li and McClelland(2022)}]{liSystematicGeneralizationEmergent2022}
Yuxuan Li and James~L. McClelland. 2022.
\newblock \href {https://doi.org/10.48550/arXiv.2210.00400} {Systematic {Generalization} and {Emergent} {Structures} in {Transformers} {Trained} on {Structured} {Tasks}}.
\newblock \emph{arXiv preprint}.
\newblock ArXiv:2210.00400 [cs].

\bibitem[{Liao et~al.(2023)Liao, Wei, Jiang, Zhang, and Ishibuchi}]{liaoDoesContinualLearning2023}
Weiduo Liao, Ying Wei, Mingchen Jiang, Qingfu Zhang, and Hisao Ishibuchi. 2023.
\newblock \href {https://proceedings.neurips.cc/paper_files/paper/2023/hash/6a42b45af2b72e6e5b5e3a6fe695809f-Abstract-Datasets_and_Benchmarks.html} {Does {Continual} {Learning} {Meet} {Compositionality}? {New} {Benchmarks} and {An} {Evaluation} {Framework}}.
\newblock \emph{Advances in Neural Information Processing Systems}, 36:33499--33513.

\bibitem[{Locatello et~al.(2020)Locatello, Weissenborn, Unterthiner, Mahendran, Heigold, Uszkoreit, Dosovitskiy, and Kipf}]{locatello2020object}
Francesco Locatello, Dirk Weissenborn, Thomas Unterthiner, Aravindh Mahendran, Georg Heigold, Jakob Uszkoreit, Alexey Dosovitskiy, and Thomas Kipf. 2020.
\newblock Object-centric learning with slot attention.
\newblock \emph{Advances in neural information processing systems}, 33:11525--11538.

\bibitem[{Luce(1996)}]{luce1996ongoing}
R~Duncan Luce. 1996.
\newblock The ongoing dialog between empirical science and measurement theory.
\newblock \emph{journal of mathematical psychology}, 40(1):78--98.

\bibitem[{Ma et~al.(2023)Ma, Hong, Gul, Gandhi, Gao, and Krishna}]{maCREPECanVisionLanguage2023}
Zixian Ma, Jerry Hong, Mustafa~Omer Gul, Mona Gandhi, Irena Gao, and Ranjay Krishna. 2023.
\newblock \href {https://doi.org/10.1109/CVPR52729.2023.01050} {{CREPE}: {Can} {Vision}-{Language} {Foundation} {Models} {Reason} {Compositionally}?}
\newblock In \emph{2023 {IEEE}/{CVF} {Conference} on {Computer} {Vision} and {Pattern} {Recognition} ({CVPR})}, pages 10910--10921, Vancouver, BC, Canada. IEEE.

\bibitem[{Mao et~al.(2019)Mao, Gan, Kohli, Tenenbaum, and Wu}]{maoNeuroSymbolicConceptLearner2019}
Jiayuan Mao, Chuang Gan, Pushmeet Kohli, Joshua~B. Tenenbaum, and Jiajun Wu. 2019.
\newblock \href {https://doi.org/10.48550/arXiv.1904.12584} {The {Neuro}-{Symbolic} {Concept} {Learner}: {Interpreting} {Scenes}, {Words}, and {Sentences} {From} {Natural} {Supervision}}.
\newblock \emph{arXiv preprint}.
\newblock ArXiv:1904.12584 [cs].

\bibitem[{McCoy et~al.(2020)McCoy, Linzen, Dunbar, and Smolensky}]{mccoy_tensor_2020}
R.~Thomas McCoy, Tal Linzen, Ewan Dunbar, and Paul Smolensky. 2020.
\newblock \href {https://aclanthology.org/2020.scil-1.34} {Tensor {Product} {Decomposition} {Networks}: {Uncovering} {Representations} of {Structure} {Learned} by {Neural} {Networks}}.
\newblock In \emph{Proceedings of the {Society} for {Computation} in {Linguistics} 2020}, pages 277--278, New York, New York. Association for Computational Linguistics.

\bibitem[{McCurdy et~al.(2024)McCurdy, Soulos, Smolensky, Fernandez, and Gao}]{mccurdyCompositionalBehaviorNeural2024}
Kate McCurdy, Paul Soulos, Paul Smolensky, Roland Fernandez, and Jianfeng Gao. 2024.
\newblock \href {https://doi.org/10.18653/v1/2024.emnlp-main.524} {Toward {Compositional} {Behavior} in {Neural} {Models}: {A} {Survey} of {Current} {Views}}.
\newblock In \emph{Proceedings of the 2024 {Conference} on {Empirical} {Methods} in {Natural} {Language} {Processing}}, pages 9323--9339, Miami, Florida, USA. Association for Computational Linguistics.

\bibitem[{Merullo et~al.(2024)Merullo, Eickhoff, and Pavlick}]{merulloLanguageModelsImplement2024}
Jack Merullo, Carsten Eickhoff, and Ellie Pavlick. 2024.
\newblock \href {https://doi.org/10.18653/v1/2024.naacl-long.281} {Language {Models} {Implement} {Simple} {Word2Vec}-style {Vector} {Arithmetic}}.
\newblock In \emph{Proceedings of the 2024 {Conference} of the {North} {American} {Chapter} of the {Association} for {Computational} {Linguistics}: {Human} {Language} {Technologies} ({Volume} 1: {Long} {Papers})}, pages 5030--5047, Mexico City, Mexico. Association for Computational Linguistics.

\bibitem[{Moskvichev et~al.(2023)Moskvichev, Odouard, and Mitchell}]{moskvichev2023conceptarc}
Arseny Moskvichev, Victor~Vikram Odouard, and Melanie Mitchell. 2023.
\newblock The conceptarc benchmark: Evaluating understanding and generalization in the arc domain.
\newblock \emph{arXiv preprint arXiv:2305.07141}.

\bibitem[{Nanda et~al.(2023)Nanda, Lee, and Wattenberg}]{nandaEmergentLinearRepresentations2023}
Neel Nanda, Andrew Lee, and Martin Wattenberg. 2023.
\newblock \href {https://doi.org/10.48550/arXiv.2309.00941} {Emergent {Linear} {Representations} in {World} {Models} of {Self}-{Supervised} {Sequence} {Models}}.
\newblock \emph{arXiv preprint}.
\newblock ArXiv:2309.00941 [cs].

\bibitem[{{nostalgebraist}(2020)}]{nostalgebraist_interpreting_2020}
{nostalgebraist}. 2020.
\newblock \href {https://www.lesswrong.com/posts/AcKRB8wDpdaN6v6ru/interpreting-gpt-the-logit-lens} {interpreting {GPT}: the logit lens}.

\bibitem[{Odouard and Mitchell(2022)}]{odouard2022evaluating}
Victor~Vikram Odouard and Melanie Mitchell. 2022.
\newblock Evaluating understanding on conceptual abstraction benchmarks.
\newblock \emph{arXiv preprint arXiv:2206.14187}.

\bibitem[{Okawa et~al.(2023)Okawa, Lubana, Dick, and Tanaka}]{okawaCompositionalAbilitiesEmerge2023}
Maya Okawa, Ekdeep~Singh Lubana, Robert~P. Dick, and Hidenori Tanaka. 2023.
\newblock \href {https://openreview.net/forum?id=ZXH8KUgFx3#all} {Compositional {Abilities} {Emerge} {Multiplicatively}: {Exploring} {Diffusion} {Models} on a {Synthetic} {Task}}.

\bibitem[{Olah(2023)}]{olahDistributedRepresentationsComposition2023}
Chris Olah. 2023.
\newblock \href {https://transformer-circuits.pub/2023/superposition-composition/index.html} {Distributed {Representations}: {Composition} \& {Superposition}}.

\bibitem[{Opiełka et~al.(2025)Opiełka, Rosenbusch, and Stevenson}]{opielkaAnalogicalReasoningLarge2025}
Gustaw Opiełka, Hannes Rosenbusch, and Claire~E. Stevenson. 2025.
\newblock \href {https://doi.org/10.48550/arXiv.2503.03666} {Analogical {Reasoning} {Inside} {Large} {Language} {Models}: {Concept} {Vectors} and the {Limits} of {Abstraction}}.
\newblock \emph{arXiv preprint}.
\newblock ArXiv:2503.03666 [cs].

\bibitem[{Park et~al.(2024)Park, Choe, and Veitch}]{parkLinearRepresentationHypothesis2024}
Kiho Park, Yo~Joong Choe, and Victor Veitch. 2024.
\newblock The linear representation hypothesis and the geometry of large language models.
\newblock In \emph{Proceedings of the 41st {International} {Conference} on {Machine} {Learning}}, volume 235 of \emph{{ICML}'24}, pages 39643--39666, Vienna, Austria. JMLR.org.

\bibitem[{Partee(2004)}]{parteeCompositionality2004}
Barbara~H. Partee. 2004.
\newblock \href {https://doi.org/10.1002/9780470751305.ch7} {Compositionality}.
\newblock In \emph{Compositionality in {Formal} {Semantics}}, pages 153--181. John Wiley \& Sons, Ltd.
\newblock Section: 7 \_eprint: https://onlinelibrary.wiley.com/doi/pdf/10.1002/9780470751305.ch7.

\bibitem[{Pavlick(2023)}]{pavlickSymbolsGroundingLarge2023}
Ellie Pavlick. 2023.
\newblock \href {https://doi.org/10.1098/rsta.2022.0041} {Symbols and grounding in large language models}.
\newblock \emph{Philosophical Transactions of the Royal Society A: Mathematical, Physical and Engineering Sciences}, 381(2251):20220041.
\newblock Publisher: Royal Society.

\bibitem[{Piaget(2013)}]{piagetConstructionRealityChild2013}
Jean Piaget. 2013.
\newblock \href {https://doi.org/10.4324/9781315009650} {\emph{The {Construction} {Of} {Reality} {In} {The} {Child}}}.
\newblock Routledge, London.

\bibitem[{Premack and Woodruff(1978)}]{premackDoesChimpanzeeHave1978}
David Premack and Guy Woodruff. 1978.
\newblock \href {https://doi.org/10.1017/S0140525X00076512} {Does the chimpanzee have a theory of mind?}
\newblock \emph{Behavioral and Brain Sciences}, 1(4):515--526.

\bibitem[{Quilty-Dunn et~al.(2023)Quilty-Dunn, Porot, and Mandelbaum}]{quilty-dunnBestGameTown2023}
Jake Quilty-Dunn, Nicolas Porot, and Eric Mandelbaum. 2023.
\newblock \href {https://doi.org/10.1017/S0140525X22002849} {The best game in town: {The} reemergence of the language-of-thought hypothesis across the cognitive sciences}.
\newblock \emph{Behavioral and Brain Sciences}, 46:e261.

\bibitem[{Ravichander et~al.(2021)Ravichander, Belinkov, and Hovy}]{ravichanderProbingProbingParadigm2021}
Abhilasha Ravichander, Yonatan Belinkov, and Eduard Hovy. 2021.
\newblock \href {https://doi.org/10.48550/arXiv.2005.00719} {Probing the {Probing} {Paradigm}: {Does} {Probing} {Accuracy} {Entail} {Task} {Relevance}?}
\newblock \emph{arXiv preprint}.
\newblock ArXiv:2005.00719 [cs].

\bibitem[{Russin et~al.(2024)Russin, McGrath, Williams, and Elber-Dorozko}]{russinFregeChatGPTCompositionality2024}
Jacob Russin, Sam~Whitman McGrath, Danielle~J. Williams, and Lotem Elber-Dorozko. 2024.
\newblock \href {https://doi.org/10.48550/arXiv.2405.15164} {From {Frege} to {chatGPT}: {Compositionality} in language, cognition, and deep neural networks}.
\newblock \emph{arXiv preprint}.
\newblock ArXiv:2405.15164 [cs].

\bibitem[{Schott et~al.(2021)Schott, Kügelgen, Träuble, Gehler, Russell, Bethge, Schölkopf, Locatello, and Brendel}]{schottVisualRepresentationLearning2021}
Lukas Schott, Julius~Von Kügelgen, Frederik Träuble, Peter~Vincent Gehler, Chris Russell, Matthias Bethge, Bernhard Schölkopf, Francesco Locatello, and Wieland Brendel. 2021.
\newblock \href {https://openreview.net/forum?id=9RUHPlladgh} {Visual {Representation} {Learning} {Does} {Not} {Generalize} {Strongly} {Within} the {Same} {Domain}}.

\bibitem[{Sen et~al.(2022)Sen, de~Carvalho, Riegel, and Gray}]{sen2022neuro}
Prithviraj Sen, Breno~WSR de~Carvalho, Ryan Riegel, and Alexander Gray. 2022.
\newblock Neuro-symbolic inductive logic programming with logical neural networks.
\newblock In \emph{Proceedings of the AAAI conference on artificial intelligence}, volume~36, pages 8212--8219.

\bibitem[{Sharkey et~al.(2025)Sharkey, Chughtai, Batson, Lindsey, Wu, Bushnaq, Goldowsky-Dill, Heimersheim, Ortega, Bloom, Biderman, Garriga-Alonso, Conmy, Nanda, Rumbelow, Wattenberg, Schoots, Miller, Michaud, Casper, Tegmark, Saunders, Bau, Todd, Geiger, Geva, Hoogland, Murfet, and McGrath}]{sharkeyOpenProblemsMechanistic2025}
Lee Sharkey, Bilal Chughtai, Joshua Batson, Jack Lindsey, Jeff Wu, Lucius Bushnaq, Nicholas Goldowsky-Dill, Stefan Heimersheim, Alejandro Ortega, Joseph Bloom, Stella Biderman, Adria Garriga-Alonso, Arthur Conmy, Neel Nanda, Jessica Rumbelow, Martin Wattenberg, Nandi Schoots, Joseph Miller, Eric~J. Michaud, Stephen Casper, Max Tegmark, William Saunders, David Bau, Eric Todd, Atticus Geiger, Mor Geva, Jesse Hoogland, Daniel Murfet, and Tom McGrath. 2025.
\newblock \href {https://doi.org/10.48550/arXiv.2501.16496} {Open {Problems} in {Mechanistic} {Interpretability}}.
\newblock \emph{arXiv preprint}.
\newblock ArXiv:2501.16496 [cs].

\bibitem[{Soulos et~al.(2024)Soulos, Conklin, Opper, Smolensky, Gao, and Fernandez}]{soulos2024compositional}
Paul Soulos, Henry Conklin, Mattia Opper, Paul Smolensky, Jianfeng Gao, and Roland Fernandez. 2024.
\newblock Compositional generalization across distributional shifts with sparse tree operations.
\newblock \emph{arXiv preprint arXiv:2412.14076}.

\bibitem[{Steedman(2019)}]{steedmanCombinatoryCategorialGrammar2019}
Mark Steedman. 2019.
\newblock \href {https://doi.org/10.1515/9783110540253-014} {Combinatory {Categorial} {Grammar}}.
\newblock In \emph{Current {Approaches} to {Syntax}}.
\newblock Publication Title: Current Approaches to Syntax.

\bibitem[{Stevens(1946)}]{stevens1946theory}
Stanley~Smith Stevens. 1946.
\newblock On the theory of scales of measurement.
\newblock \emph{Science}, 103(2684):677--680.

\bibitem[{Sun et~al.(2023)Sun, Williams, and Hupkes}]{sunValidityEvaluationResults2023}
Kaiser Sun, Adina Williams, and Dieuwke Hupkes. 2023.
\newblock \href {https://doi.org/10.18653/v1/2023.conll-1.19} {The {Validity} of {Evaluation} {Results}: {Assessing} {Concurrence} {Across} {Compositionality} {Benchmarks}}.
\newblock In \emph{Proceedings of the 27th {Conference} on {Computational} {Natural} {Language} {Learning} ({CoNLL})}, pages 274--293, Singapore. Association for Computational Linguistics.

\bibitem[{Tenney et~al.(2019{\natexlab{a}})Tenney, Das, and Pavlick}]{tenneyBERTRediscoversClassical2019}
Ian Tenney, Dipanjan Das, and Ellie Pavlick. 2019{\natexlab{a}}.
\newblock \href {https://doi.org/10.18653/v1/P19-1452} {{BERT} {Rediscovers} the {Classical} {NLP} {Pipeline}}.
\newblock In \emph{Proceedings of the 57th {Annual} {Meeting} of the {Association} for {Computational} {Linguistics}}, pages 4593--4601, Florence, Italy. Association for Computational Linguistics.

\bibitem[{Tenney et~al.(2019{\natexlab{b}})Tenney, Xia, Chen, Wang, Poliak, McCoy, Kim, Durme, Bowman, Das, and Pavlick}]{tenneyWhatYouLearn2019}
Ian Tenney, Patrick Xia, Berlin Chen, Alex Wang, Adam Poliak, R.~Thomas McCoy, Najoung Kim, Benjamin~Van Durme, Samuel~R. Bowman, Dipanjan Das, and Ellie Pavlick. 2019{\natexlab{b}}.
\newblock \href {https://doi.org/10.48550/arXiv.1905.06316} {What do you learn from context? {Probing} for sentence structure in contextualized word representations}.
\newblock \emph{arXiv preprint}.
\newblock ArXiv:1905.06316 [cs].

\bibitem[{Thomm et~al.(2024)Thomm, Camposampiero, Terzic, Hersche, Sch{\"o}lkopf, and Rahimi}]{thomm2024limits}
Jonathan Thomm, Giacomo Camposampiero, Aleksandar Terzic, Michael Hersche, Bernhard Sch{\"o}lkopf, and Abbas Rahimi. 2024.
\newblock Limits of transformer language models on learning to compose algorithms.
\newblock In \emph{The Thirty-eighth Annual Conference on Neural Information Processing Systems}.

\bibitem[{Thompson et~al.(1997)Thompson, Oden, and Boysen}]{thompsonLanguagenaiveChimpanzeesPan1997}
Roger K.~R. Thompson, David~L. Oden, and Sarah~T. Boysen. 1997.
\newblock \href {https://doi.org/10.1037/0097-7403.23.1.31} {Language-naive chimpanzees ({Pan} troglodytes) judge relations between relations in a conceptual matching-to-sample task}.
\newblock \emph{Journal of Experimental Psychology: Animal Behavior Processes}, 23(1):31--43.
\newblock Place: US Publisher: American Psychological Association.

\bibitem[{Thrush et~al.(2022)Thrush, Jiang, Bartolo, Singh, Williams, Kiela, and Ross}]{thrushWinogroundProbingVision2022}
Tristan Thrush, Ryan Jiang, Max Bartolo, Amanpreet Singh, Adina Williams, Douwe Kiela, and Candace Ross. 2022.
\newblock \href {https://doi.org/10.1109/CVPR52688.2022.00517} {Winoground: {Probing} {Vision} and {Language} {Models} for {Visio}-{Linguistic} {Compositionality}}.
\newblock In \emph{2022 {IEEE}/{CVF} {Conference} on {Computer} {Vision} and {Pattern} {Recognition} ({CVPR})}, pages 5228--5238, New Orleans, LA, USA. IEEE.

\bibitem[{Todd et~al.(2024)Todd, Li, Sharma, Mueller, Wallace, and Bau}]{toddFunctionVectorsLarge2024}
Eric Todd, Millicent~L. Li, Arnab~Sen Sharma, Aaron Mueller, Byron~C. Wallace, and David Bau. 2024.
\newblock \href {https://doi.org/10.48550/arXiv.2310.15213} {Function {Vectors} in {Large} {Language} {Models}}.
\newblock \emph{arXiv preprint}.
\newblock ArXiv:2310.15213 [cs].

\bibitem[{Vafa et~al.(2024)Vafa, Chen, Rambachan, Kleinberg, and Mullainathan}]{vafaEvaluatingWorldModel2024}
Keyon Vafa, Justin~Y. Chen, Ashesh Rambachan, Jon Kleinberg, and Sendhil Mullainathan. 2024.
\newblock \href {https://doi.org/10.48550/arXiv.2406.03689} {Evaluating the {World} {Model} {Implicit} in a {Generative} {Model}}.
\newblock \emph{arXiv preprint}.
\newblock ArXiv:2406.03689 [cs].

\bibitem[{Wattenberg and Viégas(2024)}]{wattenbergRelationalCompositionNeural2024}
Martin Wattenberg and Fernanda~B. Viégas. 2024.
\newblock \href {https://doi.org/10.48550/arXiv.2407.14662} {Relational {Composition} in {Neural} {Networks}: {A} {Survey} and {Call} to {Action}}.
\newblock \emph{arXiv preprint}.
\newblock ArXiv:2407.14662 [cs].

\bibitem[{Westerståhl(1998)}]{westerstahlMathematicalProofsVacuity1998}
Dag Westerståhl. 1998.
\newblock \href {https://www.jstor.org/stable/25001726} {On {Mathematical} {Proofs} of the {Vacuity} of {Compositionality}}.
\newblock \emph{Linguistics and Philosophy}, 21(6):635--643.
\newblock Publisher: Springer.

\bibitem[{Wu et~al.(2023)Wu, Manning, and Potts}]{wuReCOGSHowIncidental2023}
Zhengxuan Wu, Christopher~D. Manning, and Christopher Potts. 2023.
\newblock \href {https://doi.org/10.1162/tacl_a_00623} {{ReCOGS}: {How} {Incidental} {Details} of a {Logical} {Form} {Overshadow} an {Evaluation} of {Semantic} {Interpretation}}.
\newblock \emph{Transactions of the Association for Computational Linguistics}, 11:1719--1733.

\bibitem[{Young and Wasserman(1997)}]{youngEntropyDetectionPigeons1997}
Michael~E. Young and Edward~A. Wasserman. 1997.
\newblock \href {https://doi.org/10.1037/0097-7403.23.2.157} {Entropy detection by pigeons: {Response} to mixed visual displays after same–different discrimination training}.
\newblock \emph{Journal of Experimental Psychology: Animal Behavior Processes}, 23(2):157--170.
\newblock Place: US Publisher: American Psychological Association.

\bibitem[{Young and Wasserman(2001)}]{youngEntropyVariabilityDiscrimination2001}
Michael~E. Young and Edward~A. Wasserman. 2001.
\newblock \href {https://doi.org/10.1037/0278-7393.27.1.278} {Entropy and variability discrimination}.
\newblock \emph{Journal of Experimental Psychology: Learning, Memory, and Cognition}, 27(1):278--293.
\newblock Place: US Publisher: American Psychological Association.

\bibitem[{Zerroug et~al.(2022)Zerroug, Vaishnav, Colin, Musslick, and Serre}]{zerroug2022benchmark}
Aimen Zerroug, Mohit Vaishnav, Julien Colin, Sebastian Musslick, and Thomas Serre. 2022.
\newblock A benchmark for compositional visual reasoning.
\newblock \emph{Advances in neural information processing systems}, 35:29776--29788.

\bibitem[{Zhang et~al.(2019)Zhang, Gao, Jia, Zhu, and Zhu}]{zhang2019raven}
Chi Zhang, Feng Gao, Baoxiong Jia, Yixin Zhu, and Song-Chun Zhu. 2019.
\newblock Raven: A dataset for relational and analogical visual reasoning.
\newblock In \emph{Proceedings of the IEEE/CVF conference on computer vision and pattern recognition}, pages 5317--5327.

\end{thebibliography}

\end{document}